\documentclass{article}

\usepackage{arxiv}
\usepackage[utf8]{inputenc} % allow utf-8 input
\usepackage[T1]{fontenc}    % use 8-bit T1 fonts
\usepackage{hyperref}
\usepackage{url}            % simple URL typesetting
\usepackage{booktabs}       % professional-quality tables
\usepackage{amsfonts}       % blackboard math symbols
\usepackage{nicefrac}       % compact symbols for 1/2, etc.
\usepackage{microtype}      % microtypography
\usepackage{lipsum}		% Can be removed after putting your text content
\usepackage{graphicx}
\usepackage{caption}
\usepackage{subcaption}
\usepackage{float} 
\usepackage{amsmath}
\usepackage[ruled,vlined]{algorithm2e}
\usepackage[style = authortitle, citestyle=alphabetic, bibstyle=alphabetic, sorting=nyt, backend=biber, language=english, backref=true, maxcitenames=2]{biblatex}
\usepackage{multirow}
\usepackage{rotating}

\hypersetup{
	colorlinks=false,
	pdfborder={0 0 0},
}

\title{Time Series Classification using Convolutional Neural Network on Imbalanced Datasets}

\author{ 
	Department of Mobile and Embedded Systems\\
	University of Passau\\
	Innstraße 41, 94032 Passau \\
	Syed Rawshon Jamil\\
	\texttt{jamil.0505038@gmail.com} \\
}

\renewcommand{\headeright}{}
\renewcommand{\undertitle}{}
\renewcommand{\shorttitle}{TSC using CNN on Imbalanced Datasets}

\hypersetup{
pdftitle={A template for the arxiv style},
pdfsubject={q-bio.NC, q-bio.QM},
pdfauthor={David S.~Hippocampus, Elias D.~Striatum},
pdfkeywords={First keyword, Second keyword, More},
}

\begin{document}
\maketitle

%\begin{abstract}
%	\lipsum[1]
%\end{abstract}

%\let\clearpage\relax
% !TeX spellcheck = en_US
% !TeX encoding = UTF-8
\section*{Abstract}
\underline{T}ime \underline{S}eries \underline{C}lassification (TSC) has drawn a lot of attention in literature because of its broad range of applications for different domains, such as medical data mining, weather forecasting. Although TSC algorithms are designed for balanced datasets, most real-life time series datasets are imbalanced. The Skewed distribution is a problem for time series classification both in distance-based and feature-based algorithms under the condition of poor class separability. To address the  imbalance problem, both sampling-based and algorithmic approaches are used in this paper. Different   methods significantly improve time series classification's performance on imbalanced datasets. Despite having a high imbalance ratio, the result showed that F score could be as high as 97.6\% for the simulated TwoPatterns Dataset.

% keywords ca n be removed
\section{Introduction}\label{chap:introduction}
 \underline{T}ime \underline{s}eries \underline{c}lassification (TSC) is one of the main tasks among all time series data operations,  which has a vast number of applications in our daily life. For instance, A real-time warning system based on TSC has achieved significant performance compared with traditional clinical approaches and is applied in smart hospitals.

Most of the datasets related to time series are imbalanced. However, classification algorithms are designed for balanced datasets. Classification on a skewed distribution is always challenging because it is biassed towards the majority classes while ignoring minority classes. Although miss classification of a majority class is acceptable, miss classification of a minority class is dangerous.

1NN DTW is state-of-the-art in time series classification (TSC). A distance-based method might suffer despite having high-class separability in an imbalanced dataset. There is a high probability that a lot of majority class samples surround a minority class sample, and a minority class sample will be miss-classified. Nowadays, CNN is becoming popular in time series classification because of its ability to find essential  features without human supervision. We can mitigate the problem of imbalanced dataset by modifying the loss function to obtain equal contribution from majority and minority classes in loss function.

%\subsection{Research Goal}

We did some literature review of the algorithms and techniques that address an imbalanced dataset problem. The algorithms are re-evaluated on the datasets from UCR archives by making them imbalanced. We have reported which algorithm is suitable for what amount of imbalance.

\subsection{Overview}
The rest of the paper is organized as follows. A short review of related work is described in \autoref{chap:relatedwork}. \autoref{chap:background} depicts the forms, parameters and background information of imbalanced distribution. In  \autoref{chap:dataset} seven different time series datasets from the UCR archive are described. We have illustrated methodologies to address the imbalance problem in  \autoref{chap:methods}. Our experimental report on seven real-world data sets is described in \autoref{chap:results}. Finally, the paper is concluded in \autoref{chap:conclusion}. 
\section{Related Work}\label{chap:relatedwork}

This section discusses the papers related to addressing imbalanced dataset.

\cite{class_cost_sensitive_cnn_itsc} described the difficulties of time series classification on an imbalanced dataset. The author proposed to set the learning rate based on the ratio of minority samples in each mini-batch. Another approach, namely modified loss function, was proposed, ensuring that each class's contribution is equal in mini-batch. We have adopted this paper to mitigate the imbalance problem.

\cite{class_imb_cnn} described the taxonomy and parameters of an imbalanced dataset. To address the imbalance problem, sampling-based methods were proposed.  In \autoref{chap:results}, we argue our results with the imbalance parameters. 

\cite{class_mfe_msfe} proposed two novel loss functions, Mean False Error (MFE) and Mean Square False Error (MSFE), for the imbalanced datasets, which ensured equal contribution in loss from each class in a mini-batch. Experimental results exposed that better gradient descent is possible with the proposed loss function.

\cite{class_bootstrapping} proposed an algorithmic approach, namely bootstrapping, to address imbalanced distribution by creating a balanced mini-batch. The author also prioritized recall over precision in  imbalanced dataset as false negatives are more important than false positives. Bootstrapping method is adopted in this paper to mitigate imbalanced distribution.

\cite{class_adaptive} proposed a developed version of Mean Square Error (MSE), namely Global Mean Square Error (GMSE), which ensures the contribution of loss for each class is equal, and the loss function is proportional to the class separability score. The proposed loss function is used to address the imbalanced distribution. Class separability score is used in \autoref{chap:results}, to argue the reason for poor classification.

%do correct paraphrase this
\cite{class_cost_aware} proposed a novel loss function, namely, cost-aware loss, that embeds the training stage's cost information. The author showed that the loss function could also be integrated into the pre-training stage to conduct cost-aware feature extraction more effectively. Experimental results justified the validity of the novel loss function by making existing deep learning models cost-sensitive and demonstrated that the proposed model outperforms other deep models.

\section{Background}\label{chap:background}

%theory and the cow thing
We have discussed forms and parameters of an imbalanced dataset firstly on this section. Finally, the section has discussed  class separability score and its implementation.

\begin{figure}
\centering
\begin{subfigure}[b]{0.25\textwidth}
	\centering
	\includegraphics[width=\textwidth]{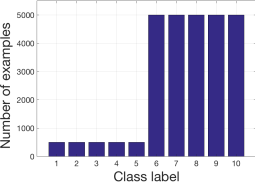}
	\caption{Step imbalance; $\rho=10, \mu=0.5$}
	\label{fig:step_imbalance_1}
\end{subfigure}
\hfill
\begin{subfigure}[b]{0.25\textwidth}
	\centering
	\includegraphics[width=\textwidth]{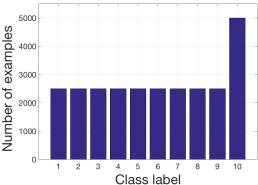}
	\caption{Step imbalance; $\rho=2, \mu=0.9$}
	\label{fig:step_imbalance_2}
\end{subfigure}
\hfill
\begin{subfigure}[b]{0.25\textwidth}
	\centering
	\includegraphics[width=\textwidth]{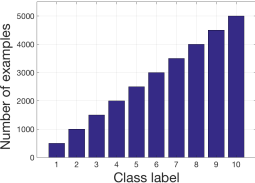}
	\caption{Linear imbalance; $\rho=10$}
	\label{fig:linear_imbalance}
\end{subfigure}
\hfill
\caption{Forms and parameters of imbalance; source \cite{class_imb_cnn}}
\label{fig:forms_of_imbalance}
\end{figure}

%\subsection{Forms and Parameters of Imbalance} 
There are two main \textit{forms of an imbalanced distribution}. Those are \cite{class_imb_cnn}:

\begin{enumerate}
\item \textit{Step imbalance} refers to a class distribution with majority class instances and minority class instances, where all the majority classes has an equal number of instances, and all the minority classes has an equal number of instances. However, the number of instances in majority classes and minority classes is not equal .  \autoref{fig:step_imbalance_1}  shows a histogram representation of a step imbalance dataset where all the majority and minority classes have 500 and 5000 instances respectively. 
%	\autoref{fig:step_imbalance_2} is also step imbalance because all the minor class has 2500 instances each and the only mojor class has 5000 instances \cite{class_imb_cnn}.

Step imbalance can be defined by two parameters as follows:

\begin{enumerate}
	\item \textit{Fraction of minority class} $\mu$ is the ratio between the number of minority classes and the total number of classes. It can be expressed by \autoref{eq:fraction_minor_class} .
	
	\begin{equation}
		\label{eq:fraction_minor_class}
		\mu = \frac{number\; of\; minority\; class}{total\; number\; of\; class}
	\end{equation}
	According to \autoref{fig:step_imbalance_1}, in a 10 class dataset,  we have 5 minority classes. So, $\mu = \frac{5}{10} = .5$. However, According to \autoref{fig:step_imbalance_2},  we have 9 minority classes out of 10 classes. So, $\mu = \frac{9}{10} = .9$ 
	
	\item \textit{Imbalance ratio} $\rho$ is the ratio between the number of instances in majority and minority classes. It can be expressed by \autoref{ir} . 
	\begin{equation}
		\label{ir}
		\rho = \frac{number\; of\; instances\; in\; majority\; class}{number\; of\; instances\; in\; minority\; class}
	\end{equation}
	According to \autoref{fig:step_imbalance_1}, majority and minority class has 5000 and 500 instances respectively. So, $\rho = \frac{5000}{500} = 10$. On the other hand, In \autoref{fig:step_imbalance_2}, majority and minority class has 5000 and 2500 instances respectively. So, $\rho = \frac{5000}{2500} = 2$ 
\end{enumerate}

\item \textit{Linear imbalance} refers to a class distribution where the number of instances of any two classes is not equal and increases gradually over classes. \autoref{fig:linear_imbalance} is a histogram representation of a linear imbalance where the number of instances increased gradually from class 1 to class 10 . 
Linear imbalance can be defined with parameter imbalance ratio  $\rho$, which is the ratio of maximum and minimum number of instances among all classes. The number of instances in intermediate classes can be interpolated. In  \autoref{fig:linear_imbalance}, we have 5000 and 500 instances in majority and minority classes respectively. So, $\rho = \frac{5000}{500} = 10$ . 
\end{enumerate}

%\subsection{Class Separability}
\textit{Class separability} is a quantitative measure that defines how well each data point falls into its own class \cite{class_adaptive}. For instance, in a binary classification problem, the class separability score for a positive instance $i$ can be defined by \autoref{class_sep} \cite{class_adaptive}. 
\begin{equation}
	\label{class_sep}
	s(i)=\frac{n(i)-p(i)}{\max \{n(i), p(i)\}}
\end{equation}
Here, $p(i)$ represents the average distance of i from all positive class instances, and $n(i)$ represents the average distance of i from all negative class instances.
% $s(i)$ can be ranged from $1$ to $-1$. The value of  $s(i) = 1$ implies the data point $i$ perfectly lies on its own class and  $-1$ implies it lies completely in opposite class.

\begin{figure}
\centering
\begin{subfigure}[b]{0.4\textwidth}
	\centering
	\includegraphics[width=\textwidth]{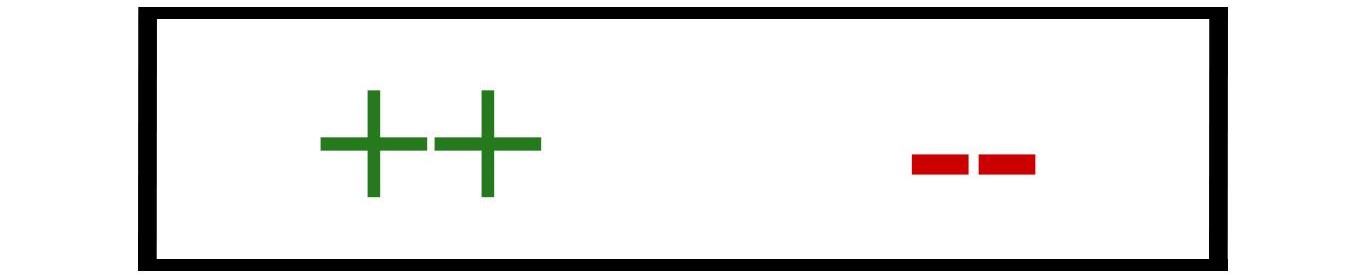}
	\caption{S $\approx$ 1}
	\label{fig:class_sep_1}
\end{subfigure}
\hfill
\begin{subfigure}[b]{0.4\textwidth}
	\centering
	\includegraphics[width=\textwidth]{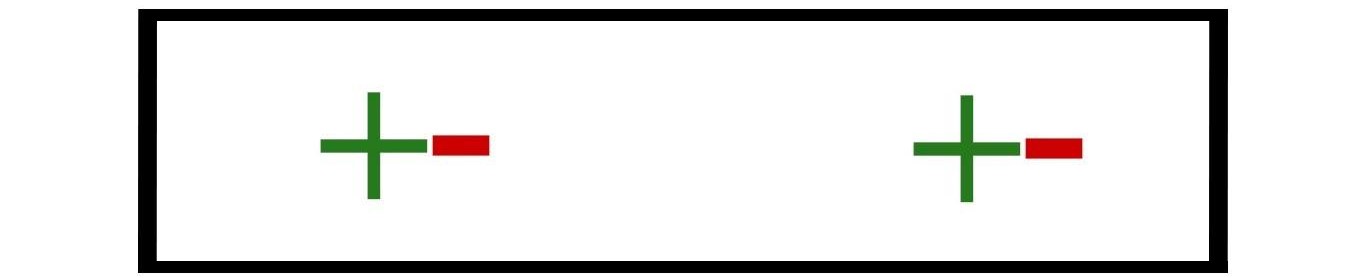}
	\caption{S $\approx$ -1}
	\label{fig:class_sep_minus_1}
\end{subfigure}

\caption{Class separability score (S)}
\label{fig:class_separability_score}
\end{figure}

The class separability score was proposed for binary classification, and adopted in  multi-class problem using binarization. The overall class separability score of a dataset is the average separability score of all instances and ranged from -1 to 1. In \autoref{fig:class_sep_1}, class separability score is nearly $1$ as positive instances and negative instances  lie with their class instances. However, the class separability score is nearly $-1$ in \autoref{fig:class_sep_minus_1} because the positive and negative instances are positioned completely with different class instances.  In  an imbalanced dataset, majority classes dominate over the score. To address this problem, majority and minority classe's contribution is considered equally by taking the average.

%hava technique ocus izmir. near to airport.

\section{Dataset}\label{chap:dataset}

%discuss cowdataset and all UCR dataset. Write exploratory data analysis here.
%take 10 dataset from the ucr dataset paper and explain why they are useful and relevant for our usecase.
%do correct give general overview of the datasets. Why there are many dataset, how you came up with that list. each individual dataset, why it is relevant. first give motivation then explain with 4-5 sentence which link to motivation.
To compare the time series classification performance, seven different datasets from the UCR archive are used in this paper \cite{UCRArchive}. Out of the seven datasets, four are EEG dataset, one is ECG dataset, one is a simulated dataset, and one is human activity recognition dataset. A summary of the datasets is given in \autoref{tab:dataset}.
\begin{table}[H]
	
	\caption{Dataset summary; source: \cite{ds_ucr} [modified]}
	\label{tab:dataset}
	%	\small
	\resizebox{\textwidth}{!}{%
		\begin{tabular}{|llllll|}
			\hline
			\hline
			Dataset               &Type & Train Cases & Dimensions & Length & \shortstack{Class separability\\ score}  \\
			ECG5000               &ECG & 500         & 1          & 140    & 0.38               \\
			TwoPatterns           &Simulated & 1000        & 1          & 128    & 0.49               \\
			HandMovementDirection &EEG & 320         & 10         & 400    & 0.0034             \\
			BasicMotions          &Human Activity Recognition & 40          & 6          & 100    & 0.54               \\
			SelfRegulationSCP1    &EEG & 268         & 6          & 896    & 0.098              \\
			SelfRegulationSCP2    &EEG & 200         & 7          & 1152   & -0.0012            \\
			
			MotorImagery          &EEG & 278         & 64         & 3000   & 0.3     \\
			\hline           		
		\end{tabular}

	}

\end{table}

\section{Methods}\label{chap:methods}

This section has discussed different methods to address imbalanced time series dataset. 

An imbalanced dataset can be addressed in two different ways as follows \cite{class_imb_cnn}:

\begin{enumerate}
	\item \textit{Data level approach}  addresses an imbalanced distribution by modifying the dataset to create a balanced distribution of the classes, and can be performed in two ways:
	\begin{enumerate}
		\item \textit{Under-sampling} removes samples from the majority classes and creates a balanced distribution for the class label. Although we are losing information, less training time is required for fewer data. Under-sampling is not suitable for a dataset with a high imbalance ratio as we are losing too much information and have very few samples to train the network. 
		
%		However, if the minority class has a large number of samples, under-sampling can be performed with a high imbalance ratio. For instance, there are 6000 positive samples and 60000 negative samples in a binary classification task. We can perform under-sampling because there are a sufficient number of samples in the minority (positive) class to train the network correctly.
		\item \textit{Oversampling} creates a balanced distribution of the class level by developing artificial instances of the minority classes.  Different oversampling approaches are given below : 
		\begin{enumerate}
			\item \textit{\underline{S}ynthetic \underline{M}inority \underline{O}ver-sampling \underline{T}echnique (SMOTE)} augments new samples by interpolating neighbors \cite{pre_data_smote}.
			
			\item \textit{Cluster-based oversampling} cluster the dataset first and then oversample each cluster separately \cite{sampling_cluster_based}.
%			\item \textit{DataBoost-IM} identifies complex examples with boosting preprocessing and uses them to generate synthetic data \cite{sampling_databoost_im}. 

		\end{enumerate}
		For a dataset with a high imbalance ratio,  oversampling will overfit because we replicate the same sample multiple times to develop artificial samples, hence losing the model’s generalization capacity.  Moreover, developing an artificial sample is a time-consuming process \cite{class_cost_sensitive_cnn_itsc}. 
	\end{enumerate}

	\item \textit{Algorithmic approach} addresses the imbalance problem by modifying the loss function  such that contribution to the loss function from majority classes and minority classes are equal. Different algorithmic approaches are given below:
	\begin{enumerate}
		\item \textit{Weighted loss in mini-batch} is an algorithmic procedure, where the contribution of loss in mini-batch from each class is considered equally 		\cite{class_cost_sensitive_cnn_itsc}. \autoref{weighted_loss} shows the loss function of this approach.
		
		\begin{eqnarray}
			\label{weighted_loss}
			E_{mini-batch}(\theta) = \dfrac{1}{|C|} \sum_{c \in All\, class} E_c(\theta)
		\end{eqnarray}
		Here, $E_c$ represents the average loss for class $c$, and $\theta$ represents the network's weights.
		
		\autoref{E_c} expresses the loss for a particular class $c$ in mini-batch.
		\begin{equation}
			\label{E_c}
			E_{c}(\boldsymbol{\theta})=\frac{1}{N_{c}} \sum_{i=1}^{N_{c}} L O S S_{i}
		\end{equation}
		Here, $N_c$ implies the number of instances in class $c$ and $LOSS_{i}$ represents categorical cross entropy loss for instance $i$.
		
		\item \textit{Bootstrapping approach} is an algorithmic approach that addresses  imbalanced dataset by creating a balanced mini-batch from the majority and minority classes \cite{class_bootstrapping}. Let us assume, in a dataset, there are  $n$, $m$ instances from majority and minority class respectively, where  $n>>m$. the batch size $s$ is selected in a way that it can be representative of all classes.
		%		 For instance, in a $4$ class problem, batch size should be $8,12,16 \, e.t.c$. 
		The number of samples in majority and minority classes per batch is defined as $s_n$ and $s_p$ respectively, where $s_{p} \approx s_{n}$ . Total number of batch per epoch $N$ is $\dfrac{n}{s_n}$ . Some negative samples might be removed from our training set in each epoch because $n$ might not be divisible by $s_n$. Those ignored samples will not create a detrimental impact on training's quality because there is a lot of samples from the majority classes.
		In each mini-batch, $s_n$ distinct majority class samples are selected. One sample from the majority class can be selected  maximum of one time per epoch. However, $s_p$ samples from the minority classes are selected randomly, and a sample from the minority class can be selected multiple times in an epoch. All the samples in the minority class have an equal probability of being chosen in each mini-batch. Such a random process guarantees that each positive instance has an equal probability of being trained with different negative instances and avoid overfitting  \cite{class_bootstrapping}. 
		\autoref{alg:bootstrapping} describes training a CNN model with bootstrapping approach.

		\begin{algorithm}[H]
			
			\caption{Train CNN with Bootstrapping \cite{class_bootstrapping}}

			\KwIn{A dataset $D$;        Number of samples from positive instances in mini-batch $s_p$; 
				\\              Number of samples from negative instances in mini-batch $s_n$;}
			\KwOut{A trained CNN model}
			
			Divide majority class samples into $N$ batches, each with $s_n$ instances\\
			$model$ $\gets$ an untrained CNN model\\
			\ForEach{epoch}
			{
				\For{$1:N$}
				{
					$batch_{majority} \gets s_{n}\: distinct\: instances\: from\: majority\: class$\\
					$batch_{minority} \gets s_{p}\: random\: instances\: from\: minority\: class$\\

					$batch_{balance} \gets batch_{majority} + batch_{minority}$\\
					Forward pass\\
					compute categorical cross entropy loss\\
					compute gradients\\
					Update weights $\theta$ of $model$ using backpropagation
					
				}

			}

			\KwRet{$model$}\;
			
			\label{alg:bootstrapping}
		\end{algorithm}

\item \textit{\underline{M}ean \underline{F}alse \underline{E}rror (MFE) and \underline{M}ean \underline{S}quare \underline{F}alse \underline{E}rror (MSFE)} are two improve loss functions of the existing \underline{M}ean \underline{S}quare \underline{E}rror  (MSE), which  address the imbalanced dataset \cite{class_mfe_msfe}.  Let us assume, in a dataset, we have total 100 samples where 10 samples are from minority class and 90 samples are from majority class. In an imbalanced dataset, minority class instances are more important than majority class instances. Therefore, minority class instances are regarded as positive instances and majority class instances are  regarded as negative instances. \autoref{tab:example-of-mfe-and-msfe} shows a confusion matrix for classification. All the loss functions are discussed based on this confusion matrix.
		%		The loss function was for binary classification and adopted in  multi-class classification using binarization. 
		
		% \usepackage{multirow}
		% \usepackage{multirow}
		\begin{table}[]
			\centering
			\caption{Confusion matrix;source:\cite{class_mfe_msfe}[modified]}
			\label{tab:example-of-mfe-and-msfe}
			\begin{tabular}{|llrrr|}
				\hline
				\hline
				
				\multicolumn{2}{|l}{\multirow{2}{*}{}} & \multicolumn{2}{l}{Prediction} &    \\				
				\multicolumn{2}{|l}{}                  & P                 & N               & \textbf{Total}   \\
				\hline
				%				\hline
				\multirow{2}{*}{\begin{turn}{90}Truth\end{turn}}    & P    & 85                & 5               & \textbf{90} \\
				& N    & 5                 & 5               & \textbf{10} \\ 
				
				\multicolumn{2}{|l}{\textbf{Total}}                  & \textbf{90}                & \textbf{10}               &   \textbf{100}\\
				\hline
				
			\end{tabular}
			
		\end{table}
		% \usepackage{multirow}
		%\begin{table}[]
		%	\begin{tabular}{lllll}
		%		\multicolumn{2}{l}{\multirow{2}{*}{}} & \multicolumn{2}{l}{Predicted Class} &    \\
		%		\multicolumn{2}{l}{}                  & P                 & N               &    \\
		%		\multirow{2}{*}{True Class}    & P    & 86                & 4               & 90 \\
		%		& N    & 5                 & 5               & 10 \\
		%		\multicolumn{2}{l}{}                  & 91                & 9               &   
		%	\end{tabular}
		%	\caption{Confusion matrix for binary classification; source: }
		%	\label{tab:example-of-mfe-and-msfe}
		%\end{table}

		\cite{class_mfe_msfe} discussed 3 different loss function as follows:
		\begin{enumerate}
			
			\item \textit{MSE} reduces the square error between prediction and ground truth and can be expressed using \autoref{mse} \cite{class_mfe_msfe}.
			\begin{equation}
				\label{mse}
				l=\frac{1}{M} \sum_{i} \sum_{n} \frac{1}{2}\left(d_{n}^{(i)}-y_{n}^{(i)}\right)^{2}
			\end{equation}
			Here, $M$ is the total number of samples. $d_{n}^{(i)}$ represents the
			ground truth value of $i^{th}$ sample on $n^{th}$ neuron while $y_{n}^{(i)}$ is the
			corresponding prediction. For instance, in the scenario of binary classification, if the $4^{th}$ sample belongs to the second class. But it is miss-classified as first class, then the ground-truth vector and prediction vector for this sample is 	
			$d^{(4)}=[0,1]^{T}$ and $y^{(4)}=[1,0]^{T}$ respectively. 			
			Here, $d_{1}^{(4)}=0$ and $d_{2}^{(4)}=1$ while $y_{1}^{(4)}=1$ and $y_{2}^{(4)}=0$. So the error of this sample is $1 / 2^{*}\left((0-1)^{\wedge} 2+(1-O)^{\wedge} 2\right)=1$\cite{class_mfe_msfe}. 
			
			According to the confusion matrix located in \autoref{tab:example-of-mfe-and-msfe} , $				l_{MSE} = \dfrac{1}{100}(5+5) = .1$ 
%			is as follows \cite{class_mfe_msfe}:
%			\begin{equation}
%				l_{MSE} = \dfrac{1}{100}(5+5) = .1
%			\end{equation}

			%			For instance, in the scenario of binary classification, if the $4^{th}$ sample actually belonged to the second class while it is predicted as the first class incorrectly,
			%			then the label vector and prediction vector for this sample is $\boldsymbol{d}^{(4)}=[0,1]^{T}$ and $\boldsymbol{y}^{(4)}=[1,0]^{T}$ respectively. We have $d_{1}^{(4)} = 0$ and $d_{2}^{(4)} = 1$ while $y_{1}^{(4)} = 1$  $y_{2}^{(4)} = 0$. So,  $ mse = 1 / 2 *\left((0-1)^{\wedge} 2+(1-0)^{\wedge} 2\right)=1$ \cite{class_mfe_msfe}

			\item \textit{Mean False Error (MFE)} is the summation of mean false positive error (FPE) and mean false negative error (FNE) \cite{class_mfe_msfe}. FPE and FNE capture errors from negative and positive class respectively. \autoref{mfe} defines MFE \cite{class_mfe_msfe}.
			\begin{equation}
				\label{mfe}
				l^{\prime}=F P E+F N E
			\end{equation}	
			and 
			\begin{equation}
				\begin{array}{c}
					F P E=\frac{1}{N} \sum_{i=1}^{N} \sum_{n} \frac{1}{2}\left(d_{n}^{(i)}-y_{n}^{(i)}\right)^{2} \\
					F N E=\frac{1}{P} \sum_{i=1}^{P} \sum_{n} \frac{1}{2}\left(d_{n}^{(i)}-y_{n}^{(i)}\right)^{2} 
				\end{array}
			\end{equation}
			$N$ and $P$ represent the numbers of instances in negative class and positive class respectively. According to the confusion matrix located in \autoref{tab:example-of-mfe-and-msfe}, $l^{\prime}=\frac{5}{10}+\frac{5}{90}=0.55$ .

			\item \textit{MSFE} is an improved version of MFE  and can be expressed by \autoref{msfe} \cite{class_mfe_msfe}.
			\begin{equation}
				\label{msfe}
				l^{\prime \prime}=F P E^{2}+F N E^{2}
			\end{equation}
			According to the confusion matrix located in \autoref{tab:example-of-mfe-and-msfe}, $l^{\prime \prime}=\left(\frac{5}{10}\right)^{2}+\left(\frac{5}{90}\right)^{2}=0.25$.
			
		\end{enumerate}
		
		Minimization of MFE implies minimization of the sum of  False Positive Error (FPE) and False Negative Error(FNE)  \cite{class_mfe_msfe}. FPE contributes more in an imbalanced distribution because the number of negative samples (majority samples) is much higher than the number of positive samples(minority samples). For instance, in the confusion matrix, we have only 10 positive samples whereas 90 negative samples. To achieve higher performance in a positive class, FNE should be pretty low. We want a high performance on difficult calf birth in our scenario. MFE is not sensitive to the positive class (minority class). MSFE solves the problem effectively \cite{class_mfe_msfe}. 
		\autoref{MSFE_BETTER} expresses MSFE \cite{class_mfe_msfe}.
		\begin{equation}
			\label{MSFE_BETTER}
			\begin{array}{l}
				MSFE=F P E^{2}+F N E^{2} \\
				\quad=\frac{1}{2}\left((F P E+F N E)^{2}+(F P E-F N E)^{2}\right)
		\end{array}\end{equation}
		The minimization operation of MSFE can find
		a minimal sum of FPE and FNE and a minimal difference. In other words, it reduces error from positive and negative classes simultaneously \cite{class_mfe_msfe}.

		\item \textit{\underline{G}lobal \underline{M}ean \underline{S}quare \underline{E}rror (GMSE)} is an improved version of Mean Square Error (MSE), which considers loss equally from the majority and minority classes and the loss can be defined with \autoref{adaptive_loss} \cite{class_adaptive}. 
		
				\begin{equation}
			\label{adaptive_loss}
			E=\frac{1}{n} \sum_{p=1}^{n} \kappa * \left(d_{p}- y_{p}\right)^{2} \text { where } \kappa=\left\{\begin{array}{ll}
				1, & \text { if } p \in \text { majority } \\
				k^{*}, & \text {otherwise }
			\end{array}\right.\end{equation}

		Here $d_{p}$ and $y_{p}$ represent the categorical value of ground truth and predicted output of $p^{th}$ instance respectively, and $n$ is the total number of training samples. 		$k^{*}$ depends on class separability score, evaluation metrics and imbalance ratio. Instead of punishing the network equally, GMSE imposes punishment based on class separability. The goal of this algorithm is to learn network weights and $k^*$ jointly. Network weights are updated after each mini-batch, whereas $k^*$ are updated after each epochs. $k^*$ is updated as follows \cite{class_adaptive}:
		\begin{equation}\kappa^{*}=\operatorname{argmin} F(\kappa) ; \quad F(\kappa)=\|T-\kappa\|^{2}\end{equation}
		and the gradient descent can be expressed as follows \cite{class_adaptive}:
		\begin{equation}
			\label{eqgradientk}
			\nabla F(\kappa)=\nabla\|T-\kappa\|^{2}=-(T-\kappa)
		\end{equation}
		Three variants of T is given below \cite{class_adaptive}:
		
		\autoref{eq_T1} \cite{class_adaptive} expresses T, to optimize G-Mean and Accuracy jointly. 
		\begin{equation}
			\label{eq_T1}
			T_{1}=H * \exp \left(-\frac{G M e a n}{2}\right) * \exp \left(-\frac{\text {Accuracy}}{2}\right)
		\end{equation}
		
		\autoref{eq_T2} \cite{class_adaptive} expresses T, to Optimize G-Mean only.
		\begin{equation}
			\label{eq_T2}
			T_{2}=H * \exp \left(-\frac{G M e a n}{2}\right)  
		\end{equation}
		
		\autoref{eq_T3} \cite{class_adaptive} expresses T, to Optimize G-Mean and validation errors $(1-Accuracy)$ jointly.
		\begin{equation}
			\label{eq_T3}
			T_{3}=H * \exp \left(-\frac{G M e a n}{2}\right) * \exp \left(-\frac{(1-A c c u r a c y)}{2}\right)
		\end{equation}
		The idea of \autoref{eq_T3} is to see whether bringing down the accuracy would help improve G-mean.

		Here $H$ represents the maximum cost for a minority class instance. $H$ depends on class separability score, imbalance ratio and can be defined as $H=I R(1+S)$.
%		\begin{equation}H=I R(1+S)\end{equation}
		Here, $IR$ represents the imbalance ratio, and $S$ represents the class separability score. 
		The value of $H$ can be ranged from $0$ to $2 \times IR$ based on class separability score. The value of $H = 2\times IR$, when the class separability score is highest $(S = 1)$ and $H = 0$, when the class separability score  is lowest $(S = -1)$. For example, GMSE punishes the network highly for miss-classifying the two patterns dataset because class separability score for this dataset is high (.54). On the other hand, it would not punish the network highly for the SelfRegulationSCP2 dataset for miss-classification for having poor class separability (-.0012). However, if the Imbalance ratio is 6 then a minority class sample contributes six times more than majority class sample to loss.
		
		\cite{class_adaptive} discovered, $T_2$ is the most suitable measure by grid search.
		
		\autoref{alg:AdaptiveGMSE} describes training a CNN model with GMSE loss.
		
		\begin{algorithm}[H]
			
			\caption{GMSE Algorithm (learnable weight); Source: \cite{class_adaptive}}

			\KwIn{A dataset $D$;}
			\KwOut{A trained CNN model}

%			\KwOut{Learned parameters, $w^{*}$ and $k^{*}$}
			
			$model$ $\gets$ an untrained CNN model\\
			$k \gets 1$\\
			\ForEach{epoch}
			{
				\ForEach{mini-batch}
				{
					Forward pass\\
					Compute loss with \autoref{adaptive_loss}\\
					Calculate gradients for error\\
					Update weights $\theta$ of $model$ using backpropagation
				}
				Compute gradients for $k$ using \autoref{eqgradientk}, with $T$ either $T_1,T_2 or T_3$ \\
				Update $k$
			}
			\KwResult{model}
			\label{alg:AdaptiveGMSE}
		\end{algorithm}

		\item \textit{Adaptive learning rate} addresses the imbalanced dataset problem by changing the learning rate based on the number of minority samples in the mini-batch. It increases the learning rate if there are many minority class instances in a mini-batch \cite{class_cost_sensitive_cnn_itsc}.

	\end{enumerate}
\end{enumerate}

\section{Results}
\label{chap:results}

%Describe the experimental setup, the used datasets/parameters and the experimental results achieved

In this section we have discussed the effect of different methods to address imbalanced datasets with four fold cross-validation. Imbalance ratio of all the dataset is set to 4.
%
%F3, AUC is reported with percentage($\%$) in \autoref{tab:method_to_address_imb_f} and \autoref{tab:method_to_address_imb_auc} respectively. Training time is reported with second (s) in \autoref{tab:method_to_address_imb_training_time}. 

According to \autoref{tab:method_to_address_imb_f}, \autoref{tab:method_to_address_imb_auc},  GMSE method is outperforming on dataset HandMovementDirection, SelfRegulationSCP2 and SelfRegulationSCP2, which has poor class separability.  For instance, the class separability score for SelfRegulationSCP2 dataset is only -.00123, and our F3 and AUC are 55.65$\%$ and 60$\%$ respectively, which is  higher than other methods. This is because only GMSE method is considering class separability score while punishing the network.  However,according to \autoref{tab:method_to_address_imb_training_time} training time for the GMSE method is higher than other methods in all datasets. For instance, in SelfRegulationSCP2 dataset, the training time for GMSE method   is 127.13 seconds, whereas training time for unweighted loss, weighted loss and bootstrapping is 109,110 and 119 seconds respectively.  GMSE is taking a high training time because computing class separability score which is a time-consuming operation.

\begin{table}[]
	\caption{Performance (F3) comparison between different method to address imbalance dataset in percentage($\%$)}
	\label{tab:method_to_address_imb_f}
	\resizebox{\textwidth}{!}{%
		\begin{tabular}{|ll|llll|}
			\hline\hline
			Dataset&Class Seperability & Unweighted loss       & Weighted loss          & GMSE          & Bootstrapping     \\
			\hline
			ECG5000               & 0.386                    & 39.09 $\pm$ 0.26 & 46.15 $\pm$ 0.58  & 21.42 $\pm$ 0.58  & \textbf{51.17 $\pm$ 2.19}  \\
			TwoPatterns           & 0.49                     & 92.06 $\pm$ 3.46 & \textbf{99.29 $\pm$ 0.03}  & 72.49 $\pm$ 0.05  & 98.33 $\pm$ 0.64  \\
			HandMovementDirection & 0.0034                   & 2.85 $\pm$ 2.85  & 3.8 $\pm$ 0.07    & \textbf{25.0 $\pm$ 0.0}    & 16.51 $\pm$ 3.37  \\
			BasicMotions          & 0.54                     & 20.54 $\pm$ 3.04 & \textbf{66.97 $\pm$ 19.25} & 11.36 $\pm$ 11.36 & 50.77 $\pm$ 27.62 \\
			SelfRegulationSCP1    & 0.098                    & 49.21 $\pm$ 3.33 & 62.09 $\pm$ 1.34  & \textbf{65.88 $\pm$ 0.0}   & 56.47 $\pm$ 3.19  \\
			SelfRegulationSCP2    & -0.00123                 & 45.65 $\pm$ 0.0  & 47.26 $\pm$ 0.01  &\textbf{55.65 $\pm$ 0.0}   & 47.72 $\pm$ 2.07  \\
			MotorImagery          & 0.3                      & 45.95 $\pm$ 0.99 & 52.09 $\pm$ 1.68  & 44.3 $\pm$ 0.66   & \textbf{57.8 $\pm$ 10.51}  \\ \hline		\end{tabular}%
	}
\end{table}

\begin{table}[]
	\caption{Performance (AUC) comparison between different method to address imbalance dataset in percentage($\%$)}
	\label{tab:method_to_address_imb_auc}
	\resizebox{\textwidth}{!}{%
		\begin{tabular}{|ll|llll|}
			\hline\hline
			Dataset& Class Seperability & Unweighted loss       & Weighted loss         & GMSE         & Bootstrapping     \\ 
			\hline
			ECG5000               & 0.386                    & 68.39 $\pm$ 0.07 & 72.05 $\pm$ 0.66 & 52.12 $\pm$ 1.05 & \textbf{75.46 $\pm$ 0.43}  \\
			TwoPatterns           & 0.49                     & 94.74 $\pm$ 2.28 & \textbf{99.53 $\pm$ 0.02} & 83.21 $\pm$ 0.01 & 98.88 $\pm$ 0.42  \\
			HandMovementDirection & 0.0034                   & 49.79 $\pm$ 0.21 & 49.63 $\pm$ 1.51 & \textbf{50.0 $\pm$ 0.0}   & 49.04 $\pm$ 1.41  \\
			BasicMotions          & 0.54                     & 58.61 $\pm$ 2.5  & \textbf{83.15 $\pm$ 6.48} & 57.5 $\pm$ 7.5   & 69.14 $\pm$ 15.81 \\
			SelfRegulationSCP1    & 0.098                    & 52.84 $\pm$ 2.84 & 64.12 $\pm$ 1.57 & \textbf{70.0 $\pm$ 0.0}   & 58.92 $\pm$ 3.58  \\
			SelfRegulationSCP2    & -0.00123                 & 50.0 $\pm$ 0.0   & 49.88 $\pm$ 1.02 & \textbf{60.0 $\pm$ 0.0}   & 49.27 $\pm$ 2.01  \\
			MotorImagery          & 0.3                      & 50.59 $\pm$ 0.59 & 55.0 $\pm$ 1.27  & 48.16 $\pm$ 1.84 & \textbf{57.84 $\pm$ 10.46} 	\\
			\hline
		\end{tabular}%
	}
\end{table}

\begin{table}[]
	\caption{Training time comparison between different method to address imbalance dataset in second}
	\label{tab:method_to_address_imb_training_time}
	\resizebox{\textwidth}{!}{%
		\begin{tabular}{|ll|llll|}
			\hline\hline
			Dataset               & Class Seperability & Unweighted loss         & Weighted loss         & GMSE           & Bootstrapping      \\
			\hline
			ECG5000               & 0.386                    & 18.56 $\pm$ 4.52   & 25.65 $\pm$ 0.61   & 34.29 $\pm$ 1.35   & 26.41 $\pm$ 0.27   \\
			TwoPatterns           & 0.49                     & 32.86 $\pm$ 0.3    & 43.82 $\pm$ 3.83   & 45.38 $\pm$ 0.74   & 40.71 $\pm$ 2.13   \\
			HandMovementDirection & 0.0034                   & 48.35 $\pm$ 0.49   & 48.94 $\pm$ 1.83   & 56.69 $\pm$ 0.2    & 50.25 $\pm$ 3.11   \\
			BasicMotions          & 0.54                     & 54.33 $\pm$ 4.52   & 59.72 $\pm$ 1.51   & 71.55 $\pm$ 3.62   & 74.61 $\pm$ 2.74   \\
			SelfRegulationSCP1    & 0.098                    & 97.37 $\pm$ 0.53   & 107.53 $\pm$ 10.1  & 125.5 $\pm$ 5.94   & 116.61 $\pm$ 22.52 \\
			SelfRegulationSCP2    & -0.00123                 & 109.59 $\pm$ 5.33  & 110.22 $\pm$ 5.36  & 127.13 $\pm$ 0.51  & 119.47 $\pm$ 0.47  \\
			MotorImagery          & 0.3                      & 163.63 $\pm$ 48.53 & 274.08 $\pm$ 18.52 & 367.86 $\pm$ 58.68 & 238.47 $\pm$ 23.97\\
			\hline
		\end{tabular}%
	}
\end{table}

\section{Conclusion}\label{chap:conclusion}

%Summarize the thesis and provide an outlook on future work.  
%Consideration of CNN is 
%\begin{enumerate}
%	\item All the TS should be equal in length
%	\item High training time. 
%\end{enumerate}
We have addressed imbalanced distribution by different algorithmic approaches where we have modified the loss function.

In \textit{summary}, we can say GMSE is suitable for a dataset with poor class separability as it considers class separability score while calculating loss. However, GMSE needs high training time because we need to compute class separability score. Bootstrapping approach is less prone to overfitting and better gradient descents is possible.

In the GMSE method, we find class separability scores only at the beginning of a training. Computing class separability score at each mini-batch can improve the performance, but it is computationally demanding. Therefore, we are leaving this for future work. In this paper, all the classes from a dataset have an equal contribution to the evaluation metrics. Developing a cost-sensitive evaluation metrics based on the class label's importance can be a possible future work.
%tested with overleaf

\printbibliography

\end{document}